\documentclass[Numbered, 12pt]{sn-jnl}
\usepackage{geometry}

\usepackage{graphicx}%
\usepackage{multirow}%
\usepackage{amsmath,amssymb,amsfonts}%
\usepackage{amsthm}%
\usepackage{mathrsfs}%
\usepackage[title]{appendix}%
\usepackage{xcolor}%
\usepackage[normalem]{ulem}%
\usepackage{hyperref}
\usepackage{textcomp}%
\usepackage{manyfoot}%
\usepackage{booktabs}%
\usepackage{algorithm}%
\usepackage{algorithmicx}%
\usepackage{algpseudocode}%
\usepackage{listings}%
\usepackage{amssymb}

\theoremstyle{thmstyleone}%
\theoremstyle{thmstyletwo}%

\theoremstyle{thmstylethree}%
\begin{document}

\title[Article Title]{Predicting Dynamics of Ultra-Large Complex Systems by Inferring Governing Equations}

\author[1]{\fnm{Qi} \sur{Shao}}
\equalcont{These authors contributed equally to this work.}
\author[1]{\fnm{Duxin} \sur{Chen}}
\equalcont{These authors contributed equally to this work.}

\author[1]{\fnm{Jiawen} \sur{Chen}}
\author[2]{\fnm{Yujie} \sur{Zeng}}
\author[2]{\fnm{Athen} \sur{Ma}}

\author[1]{\fnm{Wenwu} \sur{Yu}}\email{wwyu@seu.edu.cn}
\author[3,4,5]{\fnm{Vito} \sur{Latora}}\email{v.latora@qmul.ac.uk}
\author[6, 7]{\fnm{Wei} \sur{Lin}}\email{wlin@fudan.edu.cn}

\affil[1]{\orgdiv{School of Mathematics}, \orgname{Southeast University}, \orgaddress{\city{Nanjing}, \postcode{210096},  \country{P. R. China}}}

\affil[2]{\orgdiv{School of Electronic Engineering and Computer Science}, \orgname{Queen Mary University of London}, \orgaddress{\city{London}, \postcode{E1 4NS}, \country{UK}}}

\affil[3]{School of Mathematical Sciences, Queen Mary University of London, London E1 4NS, UK}%
\affil[4]{Dipartimento di Fisica ed Astronomia, Universit\`a di Catania and INFN, 95123, Catania, Italy}
\affil[5]{Complexity Science Hub Vienna (CSHV), Vienna, Austria}

\affil[6]{\orgdiv{Research Institute of Intelligent Complex Systems and MOE Frontiers Center for Brain Science}, \orgname{Fudan University}, \orgaddress{\city{Shanghai}, \postcode{200433}, \country{P. R. China}}}

\affil[7]{\orgdiv{School of Mathematical Sciences, LMNS, and SCMS}, \orgname{Fudan University}, \orgaddress{\city{Shanghai}, \postcode{200433}, \country{P. R. China}}}

\abstract{

{Predicting the behavior of ultra-large complex systems, from climate to biological and technological networks, is a central unsolved challenge. Existing approaches face a fundamental trade-off: equation discovery methods provide interpretability but fail to scale, while neural networks scale but operate as black boxes and often lose reliability over long times. Here, we introduce the Sparse
Identification Graph Neural Network, a framework that overcome this divide by allowing to infer the governing equations of large networked systems from data. By defining symbolic discovery as edge-level information, SIGN decouples the scalability of sparse identification from network size, enabling efficient equation discovery even in large systems.
SIGN allows to study networks with over 100,000 nodes while remaining robust to noise, sparse sampling, and missing data. Across diverse benchmark systems, including coupled chaotic oscillators, neural dynamics, and epidemic spreading, it recovers governing equations with high precision and sustains accurate long-term predictions.
Applied to a data set of time series of temperature measurements in 71,987 sea surface positions, SIGN identifies a compact predictive network model and captures large-scale sea surface temperature conditions up to two years in advance. By enabling equation discovery at previously inaccessible scales, SIGN opens a path toward interpretable and reliable prediction of real-world complex systems. }
}

\maketitle 

\section{Introduction}
\noindent
The prediction of collective dynamics in ultra-large-scale networked systems has traditionally been framed as a high-dimensional prediction problem, in which expressive models are trained to map past observations to future states~\cite{Grewe_Langer_Kasper_Kampa_Helmchen_2010, Colizza_2006, chen2023tracking}. 
Despite substantial advances in data acquisition and computational capacity, this paradigm has shown limited success for systems with tens of thousands to millions of interacting components, particularly for long-horizon prediction under noise and heterogeneity~\cite{meena2023emergent, guo2023generative, Nitzan_Casadiego_Timme_2017, tiranov2023collective}. 
This suggests that the central challenge lies not in data availability or model expressiveness, but in the predictive representation itself.

Throughout, we refer to ultra-large-scale networks as systems with $N = \mathcal{O}(10^4\text{--}10^5)$ nodes (or more), where per-node model fitting becomes computationally and statistically prohibitive, and where collective dynamics are governed by shared interaction mechanisms rather than independent node-level processes~\cite{Chang_Pierson_2021}.
As a result, scalable and reliable prediction requires a compact representation that explicitly encodes these governing interaction rules and remains invariant as network size increases. 
However, explicitly enumerating such interactions becomes prohibitive at scale, as the number of candidate terms grows quadratically with the number of components, rendering conventional equation discovery and network-scale prediction computationally intractable \cite{hu2025learning}. 
This inherent structural constraint indicates that robust prediction in ultra-large-scale systems must be reformulated as an equation-inference-driven problem, in which predictive performance depends on inferring shared governing equations in a manner that is independent of network size.

Existing approaches fall broadly into two families, but each satisfies only part of this requirement.  
Symbolic sparse identification methods, such as SINDy and its networked variants~\cite{brunton2016discovering, casadiego2017model}, explicitly infer governing equations but rely on separate regressions at individual nodes, leading to poor scalability, statistical instability, and inconsistent functional forms that limit their use for network-scale prediction \cite{gao2022autonomous, gao2024learning, zolman2025sindy, yu2025discover}. 
Neural predictors, including graph neural networks and physics-informed architectures~\cite{raissi2019physics, chen2021physics}, scale efficiently to large systems but typically operate without inferring explicit equations, resulting in high-dimensional representations that generalize unreliably under noise, partial observability, or distributional shift~\cite{hamilton2017inductive, chiang2019cluster}.

As a result, a fundamental gap remains: scalable frameworks for ultra-large-scale prediction based on directly inferred governing equations of networked dynamics are still lacking.
Bridging this gap is essential for predictive modeling and control in domains such as gene regulation~\cite{wang2023dictys}, collective motion~\cite{lukeman2010inferring}, and climate diagnostics~\cite{westberry2023atmospheric}. 
To address this challenge, we introduce the \emph{Sparse Identification Graph Neural Network} (SIGN), a unified framework that enables scalable prediction through equation inference. 
SIGN reformulates equation discovery as a shared message-passing process, learning a common set of candidate self and pairwise interaction functions represented as symbolic modules and selecting the governing equations via sparsity-promoting mechanisms. The learned dynamics are trained using multi-step state rollout loss, and predictions are obtained by numerically integrating the inferred system forward in time.
Crucially, the parameter count of SIGN depends on the number of interaction types rather than the number of nodes, enabling equation-based long-horizon prediction in networks with up to $10^5$ elements.

By directly inferring governing equations, SIGN provides a compact predictive representation that generalizes beyond the training window. We validate this paradigm on six benchmark systems spanning regulation, contagion, synchronization, and neural excitability, and we further demonstrate equation-based prediction on large-scale neural and real-world geophysical data. Together, these experiments show that scalable prediction can be obtained by first recovering shared governing equations and then integrating the inferred dynamics.

\section{Results} \label{resul}
\begin{figure}[h]
    \centering
    \includegraphics[width=1\linewidth]{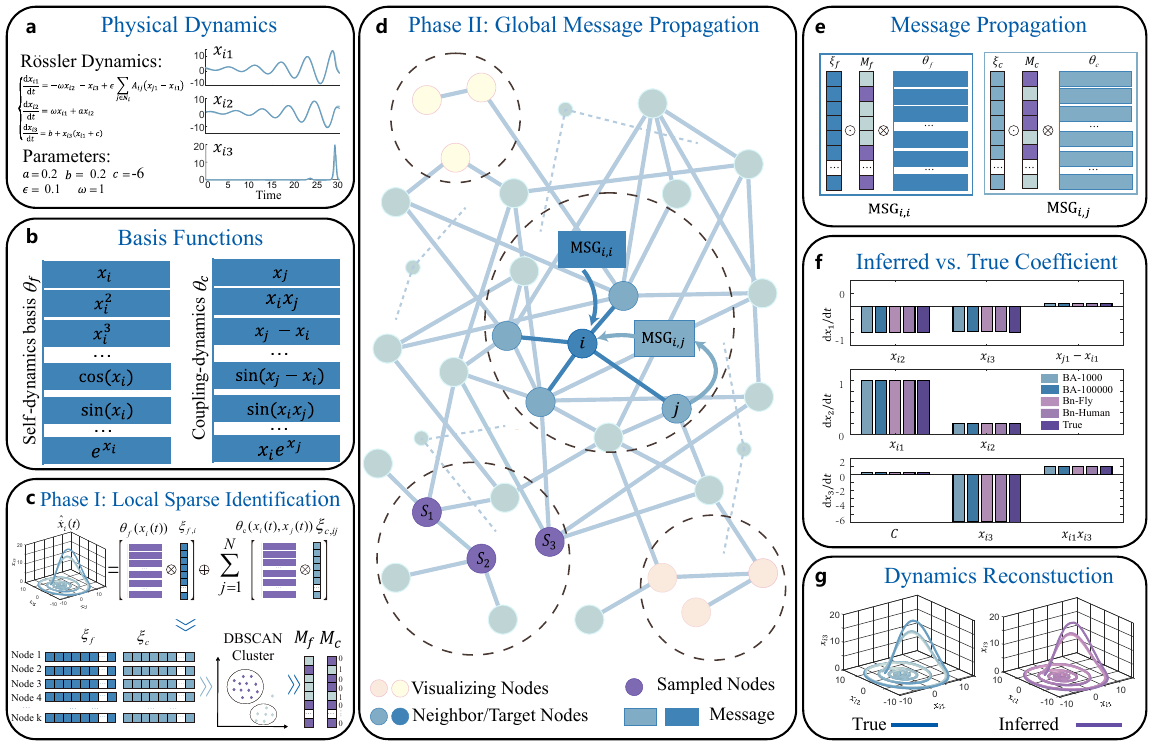}
    \caption{
\textbf{SIGN pipeline.} {SIGN facilitates accurate and scalable equation discovery through a two-phase process.} \textbf{a}, {An example of the network of coupled R\"ossler oscillators that we aim to reconstruct from its time-series observations.} \textbf{b}, Candidate libraries for intrinsic terms \(F\) and coupling terms \(C\), each generated from a set of nonlinear basis functions. \textbf{c}, Phase~I: Sparse regression on a small subset of nodes, followed by {DBSCAN clustering to determine a global support mask.} \textbf{d}, Phase~II: Message passing on the full graph using the learned mask to estimate the shared coefficients of \(F\) and \(C\). \textbf{e}, Masked intrinsic and coupling messages, used to form the recovered equations. \textbf{f}, Coefficient recovery error (sMAPE) across network topologies and sizes. \textbf{g}, {Reconstructed versus true trajectories from integrating the inferred equations, showing the model’s predictive accuracy.}
}
\label{fig1}
\end{figure}

\subsection{Decoupling identification from network scale via the SIGN framework}
\noindent
We consider networked systems whose dynamical equations can be written as~\cite{barzel2013universality}
\begin{equation}\label{eq:base}
\frac{{\rm d}x_i(t)}{{\rm d}t}
= F(x_i(t)) + \sum_{j=1}^{n} A_{ij}\, C(x_i(t), x_j(t)),
\end{equation}
where $n$ is the number of nodes, {$x_i(t)\in\mathbb{R}^d$ denotes the state of node $i$,} {$F(\cdot)$ is the intrinsic (self-dynamics) function, and $C(\cdot,\cdot)$ is the pairwise coupling function that is \emph{node-invariant} across the network.} {The adjacency matrix $A\in\mathbb{R}^{n\times n}$ specifies which node pairs interact; unless otherwise stated, we assume $A$ is known from the system topology (often sparse in large networks).} {Given time-series observations of $\{x_i(t)\}_{i=1}^n$, our goal is to recover the functional forms and coefficients of $F$ and $C$ under this node-invariant (globally shared across nodes) structure.}

A common strategy for discovering these functions is to perform sparse regression independently at each node using a library of nonlinear candidate terms.  
However, this per-node formulation faces two fundamental bottlenecks.  
First, it requires solving \(n\) regressions, causing computational cost to scale linearly with network size and making symbolic identification infeasible for large systems.  
Second, because each node observes different initial conditions and neighborhood configurations, these independent regressions often yield inconsistent sparsity patterns, even though the true governing structure is {node-invariant.}  
These inconsistencies accumulate with increasing \(n\), severely limiting the robustness and scalability of existing sparse-identification approaches.

To overcome these challenges, we introduce SIGN, a unified two-phase framework that decouples the complexity of equation discovery from that  of the network size.  
SIGN leverages the fact that the functional form of \(F\) and \(C\) is the same across nodes: Phase~I recovers a robust global sparsity pattern, and Phase~II estimates the associated coefficients through a shared message-passing model.
This eliminates per-node regressions entirely and reduces the discovery problem to learning a small, node-count–independent parameter set.

\vspace{4pt}
\noindent\textbf{Phase I: recovering a global symbolic support.}
{After constructing the library of candidate functions to use as basis for F and C (Fig.~\ref{fig1}\textbf{b}), we sample a small subset of nodes (e.g., \(k=50\)) and perform sparse regression at each sampled node to obtain binary masks \(M_i\) that indicate active candidate terms.}
Under shared dynamics, these masks represent noisy estimates of the same underlying sparsity pattern. 
We cluster the collection \(\{M_i\}\) using DBSCAN~\cite{ester1996density} to extract a consistent consensus mask (Fig.~\ref{fig1}\textbf{c}). 
This mask defines the global candidate support and drastically reduces the search space for subsequent optimization.

\vspace{4pt}
\noindent\textbf{Phase II: learning shared coefficients via message passing.}
Given the consensus support, SIGN learns the intrinsic and coupling coefficients \(\xi_f\) and \(\xi_c\) through a GNN that propagates dynamical information across the full topology (Fig.~\ref{fig1}\textbf{d}). 
For each node \(i\),
\begin{equation}
\mathrm{MSG}_{i,i} = (\xi_f \odot M_f)^\top \theta_f(x_i), \qquad
\mathrm{MSG}_{i,j} = (\xi_c \odot M_c)^\top \theta_c(x_i, x_j),
\end{equation}
where the Hadamard product enforces the symbolic support discovered in Phase~I. 
Because the coefficients are shared across all nodes, the number of trainable parameters depends only on the number of active terms, not on \(n\), making large-scale identification computationally efficient.

{We present the full workflow of SIGN applied to a network of \(10^5\) coupled R\"{o}ssler oscillators (Fig.~\ref{fig1}). 
In Fig.~\ref{fig1}\textbf{a}, we introduce the network dynamics to be identified, followed by the corresponding candidate library in Fig.~\ref{fig1}\textbf{b}. 
Phase I (Fig.~\ref{fig1}\textbf{c}) involves performing sparse regression at each node using the candidate library, identifying the relevant terms that contribute to the governing equations of the network. This process yields a support mask that highlights the terms present in the governing dynamics. 
Phase II (Fig.~\ref{fig1}\textbf{d}) utilizes message passing across the entire graph, propagating the identified support information to learn the shared coefficients that characterize the dynamics at all nodes. Fig.~\ref{fig1}\textbf{e} depicts the form of the message passed during Phase II. 
The recovered governing equations are summarized in Fig.~\ref{fig1}\textbf{f}, while Fig.~\ref{fig1}\textbf{g} displays the reconstructed trajectories for all nodes. 
Since Phase II optimizes a fixed set of shared coefficients, the parameter count remains independent of \(n\). Empirical runtime scaling is reported in Fig.~\ref{fig3}\textbf{j}.}

By decoupling symbolic structure discovery from network scale, SIGN enables efficient and interpretable recovery of governing dynamics in systems with up to \(10^5\) nodes. 
{This decoupling is the key mechanism that enables scalable identification and, via equation-based integration, prediction in the following sections.}
Full implementation details and theoretical analysis are provided in Section~\ref{sce:method} and \textcolor{blue}{the SI, Section~I}.

\begin{figure}
    \centering
    \includegraphics[width=1\linewidth]{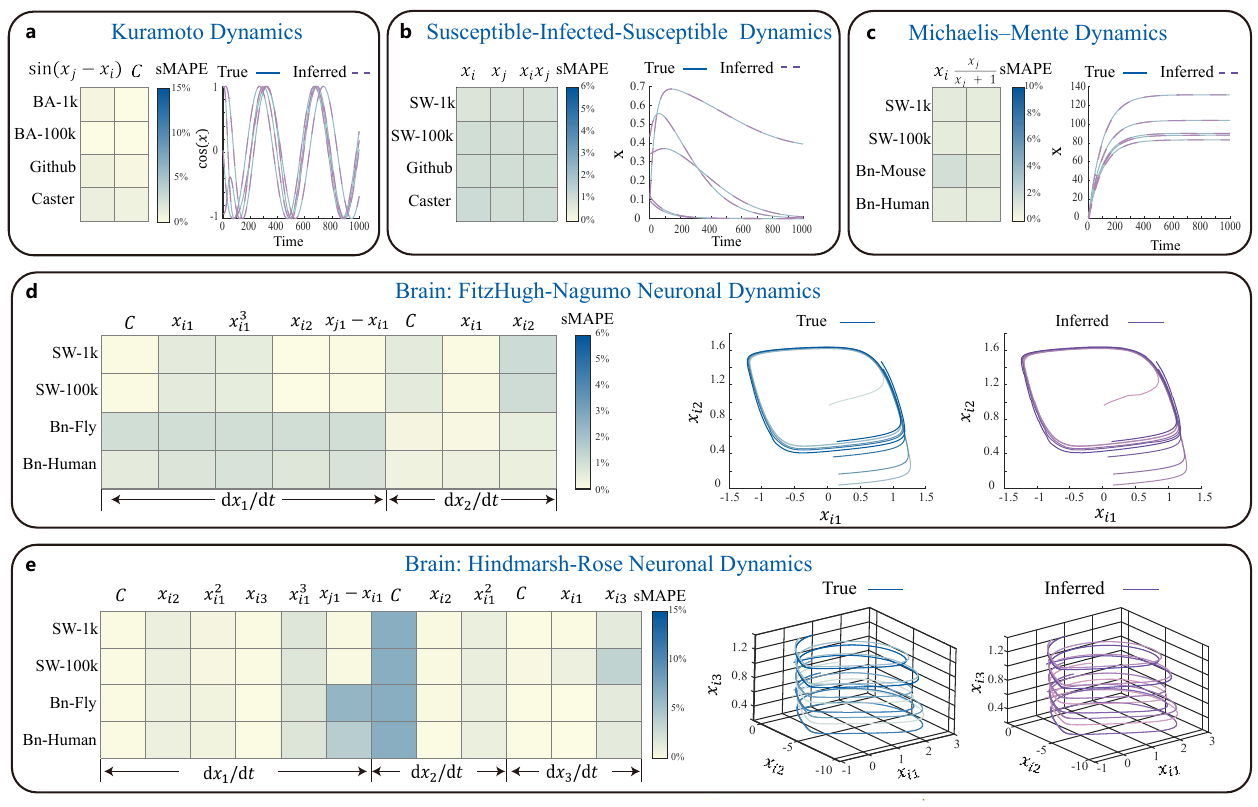}
\caption{
Equation discovery across diverse networked dynamical systems using the SIGN framework.
\textbf{a}, Kuramoto phase-oscillator network.
Here \(x_i(t)\) denotes the phase of oscillator \(i\), and the interaction between connected oscillators follows the sinusoidal coupling \(\sin(x_j-x_i)\), where \(j\) ranges over neighbors of \(i\) defined by the network adjacency \(A_{ij}\) (i.e., edges indicate which pairs interact).
The left panel reports the inference error of the coupling-term coefficient across synthetic scale-free networks with sizes of $10^3$ and $10^5$ nodes, respectively, and two large empirical networks (GitHub and Catster), while the right panel shows an example of true versus reconstructed phase trajectories \(x_i(t)\) for three representative nodes.
\textbf{b}, SIS model on small-world and empirical networks: coefficient inference errors for the epidemic-spreading dynamics across network types and sizes.
\textbf{c}, Michaelis--Menten (MM) regulatory model: coefficient inference errors for gene-regulatory dynamics across multiple network topologies and scales.
\textbf{d}, FitzHugh--Nagumo model: coefficient inference errors for synthetic simulations and a large-scale empirical brain network.
\textbf{e}, Hindmarsh--Rose model: coefficient inference errors across model terms, highlighting the distribution of errors across coefficients.}

\label{fig2}
\end{figure}
\subsection{Universal recovery of governing equations across diverse benchmarks}

\noindent
{Having introduced SIGN and illustrated the full pipeline on a representative system, namely the R\"ossler network of coupled oscillators (Fig.~\ref{fig1}), We ask whether SIGN can recover \emph{node-invariant} intrinsic and coupling laws from time series across various dynamics, network sizes, and topologies using one common inference procedure. We now evaluate whether the same equation-discovery paradigm generalizes across six canonical networked dynamical processes: We can solve}
Michaelis--Menten (MM) regulation~\cite{mazur2009reconstructing}, Susceptible--Infected--Susceptible (SIS) epidemic spreading~\cite{pastor2015epidemic}, Kuramoto synchronization~\cite{strogatz2001exploring}, coupled R\"ossler oscillators~\cite{barahona2002synchronization}, the FitzHugh--Nagumo (FHN) model~\cite{rabinovich2006dynamical}, and the Hindmarsh--Rose (HR) neuronal system~\cite{rabinovich2006dynamical}.  
{These benchmarks span biochemical regulation, contagion-like spreading, phase synchronization, and low- to high-dimensional neural excitability, allowing us to test both generalizability (various network structures and dynamics) and scalability (robust inference as the network grows).}

\medskip
For each system, we fit SIGN from time-series observations using a fixed candidate library and report coefficient errors via sMAPE. We then evaluate prediction by numerically integrating the inferred equations from held-out initial conditions, without introducing an additional predictive model. Specifically, we use data from 1000 time steps of the dynamics generated by all nodes in the specific network as observational data, and apply the SIGN model to infer the possible dynamical equations based on the entire dataset. All reconstructed trajectories are predicted using the inferred dynamics and initial states. A prediction example, can be found in Section~\ref{main:24}. We assess the accuracy of the inferred equations using sMAPE, where the definition of sMAPE is provided in Section~\ref{main:metr}.

\medskip
\noindent
\textbf{Low-dimensional nonlinear systems: Kuramoto, SIS, and MM} (Fig.~\ref{fig2}\textbf{a--c}).
{We first consider three canonical low-dimensional nonlinear dynamics where the node's state space consists of a single variable. We vary both network topology and scale,} including synthetic Barab\'asi--Albert (BA) and Watts--Strogatz (small-world) graphs as well as large empirical networks. 
Fig.~\ref{fig2}\textbf{a} reports Kuramoto coupling recovery on scale-free networks across increasing sizes and on two empirical interaction graphs (e.g. Catster network with 149k nodes and 5.4M edges).
Fig.~\ref{fig2}\textbf{b} reports SIS recovery on small-world and empirical networks.
Fig.~\ref{fig2}\textbf{c} reports MM regulation recovery on gene-regulatory networks across multiple scales.
Across all settings, 
{the inferred intrinsic and coupling terms closely match the ground truth, with coefficient errors below \(1\%\) across network sizes and topologies.}
{Using the recovered equations, we can integrate forward from held-out initial conditions to obtain predicts, using the same inferred equation form across network sizes.}

\medskip
\noindent
{\textbf{High-dimensional neural dynamics: FHN and HR} (Fig.~\ref{fig2}\textbf{d--e}).}
{Next, we assess SIGN on two higher-dimensional neuronal systems where the node's state space involves multiple variables.}
{For the two-dimensional FitzHugh--Nagumo model, we evaluate recovery on both synthetic small-world networks and a large empirical human-brain network  (88k nodes), observing consistently low coefficient errors across settings. The human-brain network is sourced from BigBrain \cite{bigbrain} and partitioned using the METIS algorithm \cite{karypis1998fast}. In this section, the network is divided into two segments, while in Section~\ref{main:24}, it is divided into eight segments (Fig.~\ref{fig2}\textbf{d}).}
{For the three-dimensional Hindmarsh--Rose system (Fig.~\ref{fig2}\textbf{e}), recovery remains stable across coefficients despite the strongly nonlinear and chaotic dynamics.}
{Across both fly-brain (2k nodes) and large-scale human-brain networks, integrating the inferred equations reproduces the observed trajectories, supporting equation-based prediction for multi-dimensional nonlinear oscillators.}

\medskip
{Overall, across six diverse dynamical systems, SIGN recovers explicit governing equations on networks up to \(10^5\) nodes and up to \(10^7\) edges.}
{Crucially for our central thesis, once explicit equations are recovered, prediction follows by direct numerical integration—providing a scalable prediction route that remains interpretable and does not require retraining a separate predictor for each network size.}
Detailed error analyses and simulation settings are provided in 
\textcolor{blue}{the SI, Section~IV}.

\begin{figure}
    \centering
    \includegraphics[width=0.95\linewidth]{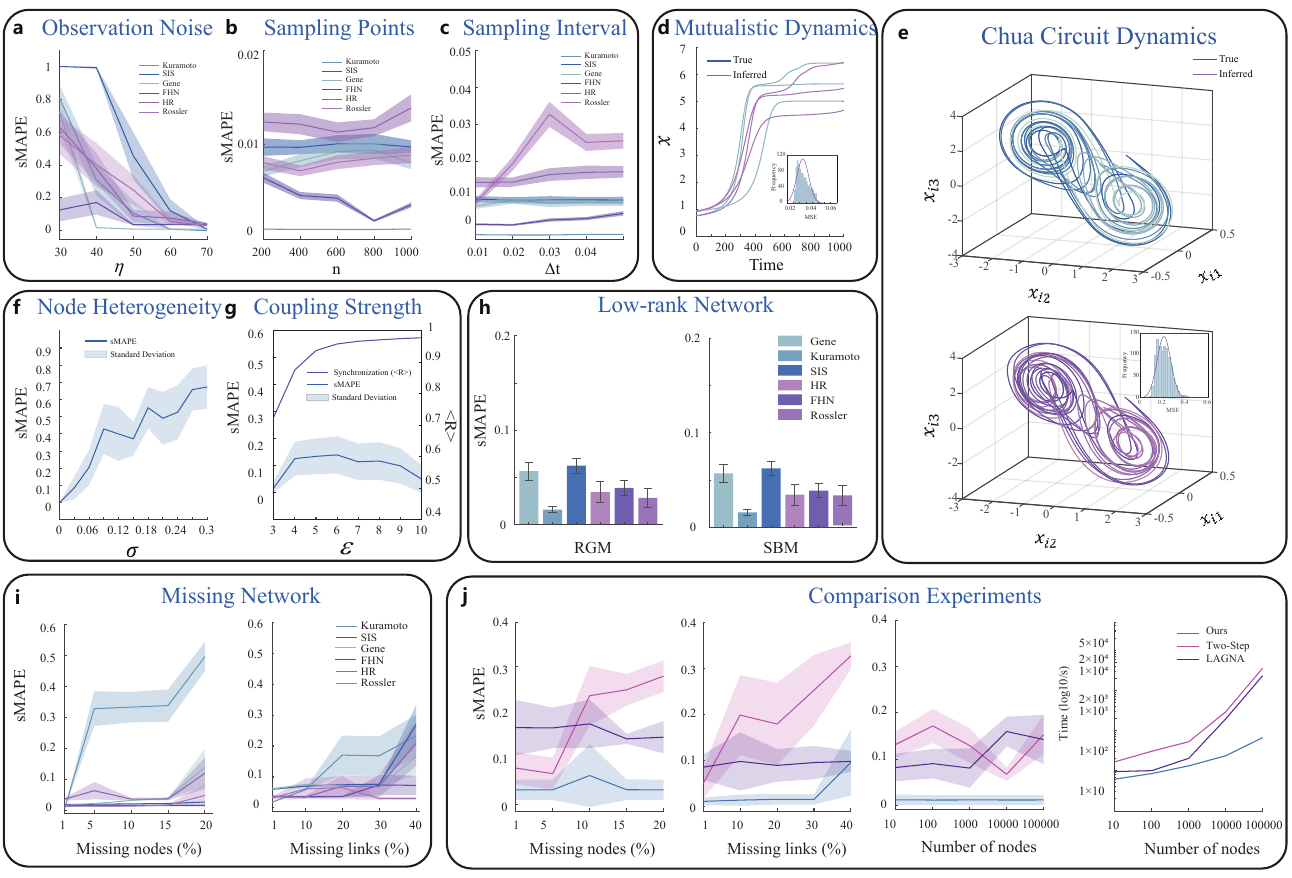}
\caption{
SIGN supports equation inference under noise, limited observations, complex (out-of-basis) dynamics, and imperfect network information.
{All experiments are repeated 10 times. Panels report coefficient errors (sMAPE) and representative trajectory reconstructions where applicable.}
\textbf{a}, Noise robustness: sMAPE under additive Gaussian noise as the signal-to-noise ratio (SNR) decreases (down to 60\,dB) across the benchmark systems.
\textbf{b,c}, Sampling resolution on ultra-large networks (\(10^5\) nodes): \textbf{b} shows sMAPE as the number of observed time points is reduced (down to \( 200\)); \textbf{c} shows sMAPE as the sampling interval \(\Delta t\) is varied (with \(\Delta t=0.01 \sim 0.05\)).
\textbf{d,e}, Non-canonical dynamics (basis mismatch): representative trajectory reconstructions for systems whose generating functions are not contained in the candidate library, including a mutualistic model with fractional nonlinearities and the chaotic Chua circuit; reconstruction error is quantified by mean squared error (MSE) across nodes.
\textbf{f}, Phase heterogeneity: Kuramoto oscillators with natural frequencies drawn from \(\mathcal{N}(0.3,\sigma)\); sMAPE as the dispersion \(\sigma\) increases.
\textbf{g}, Strong coupling: FitzHugh--Nagumo networks as coupling strength increases toward synchronized dynamics, quantified by the order parameter \(\langle R\rangle \to 1\).
\textbf{h}, Structured topologies: sMAPE on low-rank random graph models (RGM) and stochastic block models (SBM) with \(1{,}000\) nodes.
\textbf{i}, Structural incompleteness: sMAPE under missingness, with up to 20\% node removal and 40\% edge deletion prior to inference.
\textbf{j}, Baselines and scalability: coefficient-error comparison with Two-stage and LaGNA under structural incompleteness, and runtime scaling with network size from \(10\) to \(10^5\) nodes.
}

\label{fig3}
\end{figure}

\subsection{Stability under noise and incomplete data}
\label{sec23}

\noindent
{We next evaluate factors affecting the robustness of equation-based prediction, where predictions are obtained by numerically integrating the inferred equations. These factors include observational noise, limited sampling resolution, basis mismatch, complex dynamics, parameter heterogeneity, strong coupling, structured topologies, and missing nodes/edges. For each factor, we perform a robustness analysis by systematically varying its level and assess its impact on the inferred equations' stability. We report coefficient errors (sMAPE) and evaluate whether the integration of the inferred equations results in stable trajectories, ensuring the prediction method is reliable under different conditions.}

\medskip
\noindent\textbf{Robustness to observational noise and limited sampling.}
We first assess the effect of observational noise by injecting additive Gaussian perturbations with signal-to-noise ratios (SNRs) from 70~dB down to 30~dB {to the simulated dynamical data generated on networks with $10^5$ nodes.} (Fig.~\ref{fig3}\textbf{a}).  
{Across the six systems, coefficient recovery degrades smoothly as SNR decreases, then remains stable at moderate noise levels.}
{Most systems maintain low coefficient error down to 60~dB; the SIS model exhibits earlier degradation, consistent with ambiguity among interchangeable terms in its functional structure.}

{We next vary sampling resolution on networks with \(10^5\) nodes: the number of observed time points (Fig.~\ref{fig3}\textbf{b}) and the sampling interval \(\Delta t\) (Fig.~\ref{fig3}\textbf{c}).}
With at least \(200\) samples at \(\Delta t = 0.01\), all systems retain sMAPE errors below \(1\%\).  
{As sampling becomes coarser, coefficient errors increase modestly, with larger sensitivity for the higher-dimensional bursting dynamics in HR.}
{Across these settings, the inferred equations remain suitable for forward integration, supporting equation-based prediction under limited temporal resolution.}

\medskip
\noindent\textbf{Robustness to intrinsic dynamical complexity.}
We further test SIGN under factors that are intrinsic to the underlying dynamics.  
For the Kuramoto system, we increase heterogeneity in natural frequencies by varying the standard deviation \(\sigma\) of the phase distribution from 0.03 to 0.3 (Fig.~\ref{fig3}\textbf{f}).  
{We found that the coefficient error increases approximately linearly with \(\sigma\), reflecting the increasing diversity of node-level trajectories under heterogeneous parameters.}

For the FitzHugh--Nagumo model, we strengthen coupling by increasing the interaction coefficient \(\epsilon\) (Fig.~\ref{fig3}\textbf{g}).  
We report coefficient error as coupling increases toward near-synchrony (\(\langle R \rangle \rightarrow 1\)), where trajectories become highly correlated and identification becomes more challenging. {However, the error does not increase and remains within a relatively low range.}

{We also consider the case of basis mismatch, where the true functional forms are not contained in the candidate library (e.g., $\frac{x_i x_j}{D_i+E_i x_i+H_j x_j}$ in mutualistic interaction model and $|x + 1| - |x - 1|$ in Chua circuit).}
In the mutualistic interaction model (fractional nonlinearities) and the Chua circuit (featuring discontinuous absolute-value nonlinearities), 
{we fit SIGN with standard polynomial and trigonometric functions and assess the trajectory reconstruction.}
{Results shown for systems of $N = 10^3$ nodes show that reconstructions remain stable for all nodes, indicating that the inferred equations can serve as predictive surrogates even under library mismatch  (Fig.~\ref{fig3}\textbf{d,e}).}

\medskip
\noindent\textbf{Robustness to structural incompleteness.}
Real-world networks frequently contain missing nodes, missing edges, or low-rank networks.  
{We first test low-rank networks using Random Geometric Models (RGM) and Stochastic Block Models (SBM) with \(1{,}000\) nodes, reporting sMAPE across the six systems (Fig.~\ref{fig3}\textbf{h}).}
On a \(10^5\)-node network, SIGN tolerates up to \(20\%\) node removal while 
{maintaining low coefficient error for five of the six systems}
(Fig.~\ref{fig3}\textbf{i}, left).  
{The Kuramoto model differnetly from the other systems shows higher sensitivity under node removal, consistent with its reliance on neighbor-phase interactions.}
Under edge removal (Fig.~\ref{fig3}\textbf{i}, right), all six systems remain robust up to \(30\%–40\%\) edge deletion, with five models retaining sMAPE below \(10\%\).  
{These perturbation tests quantify how missing information in the graph affects node-invariant dynamics recovery and the downstream stability of equation-based integration.}

\medskip
\noindent\textbf{Baseline comparison and scalability implications.}
Finally, we compare SIGN with two recent baselines, including Two-stage (TP-SINDy)~\cite{gao2022autonomous} and LaGNA~\cite{gao2024learning}, under node deletion, edge deletion, network scaling (10 to \(10^5\) nodes), and inference time (Fig.~\ref{fig3}\textbf{j}).  
{SIGN yields sMAPE $\leq 1\%$ in this setting, while TP-SINDy and LaGNA reach errors on the order of $10\%$ (Fig.~\ref{fig3}\textbf{j}). We also report wall-clock time as a function of network size; SIGN completes inference on $10^5$ nodes within a few hundred seconds in our implementation, whereas the baselines require substantially longer runtimes due to per-node regression and iterative refinement.}
\medskip

Together, these perturbation studies characterize when shared equation recovery remains stable and when it degrades, directly informing the reliability of prediction over long horizons.

\begin{figure}[t]
    \centering
    \includegraphics[width=1\linewidth]{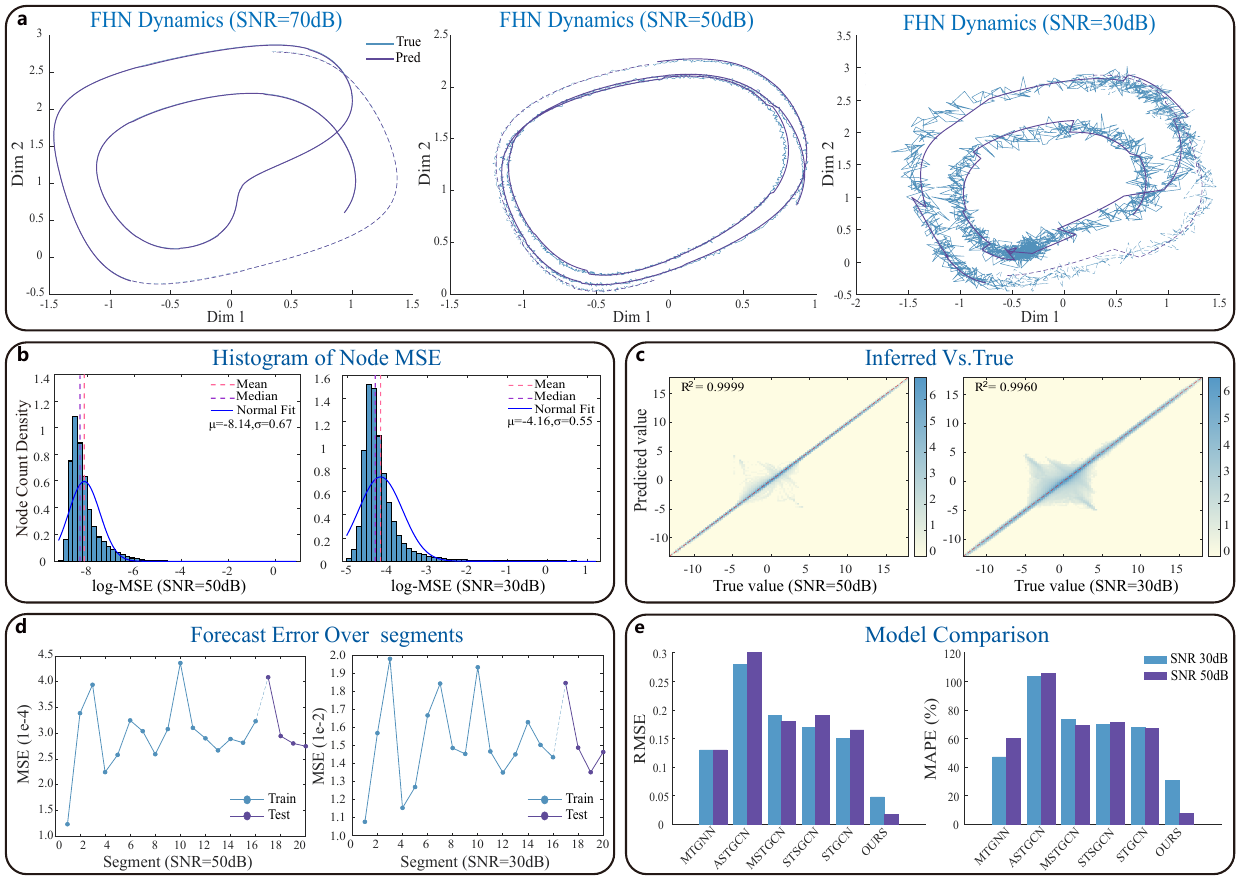}
    \caption{
Scalable prediction of large-scale FitzHugh--Nagumo neural dynamics under observational noise using SIGN.
\textbf{a}, Representative phase-space trajectories obtained by integrating the inferred equations, shown for multiple noise conditions; solid and dashed curves indicate the training and testing intervals, respectively, and nodes are randomly sampled with noise intensities centered around the median.
\textbf{b}, Distribution of node-wise prediction error (MSE) at SNR \(=50\) dB (left) and \(=30\) dB (right).
\textbf{c}, Two-dimensional density plots comparing true versus predicted values at SNR \(=50\) dB (left) and \(=30\) dB (right); density indicates how frequently pairs fall into the same region.
\textbf{d}, Prediction error (MSE) evaluated across temporal segments, with the training and testing intervals indicated.
\textbf{e}, Comparison on a \(1{,}000\)-node simulated dataset under SNR \(=50\) dB and \(=30\) dB, reporting predictive errors for SIGN and neural-network-based predictors.
}
    \label{fig41} 
\end{figure}
\subsection{Prediction of large-scale noisy neural dynamics}
\label{main:24}
\noindent
Finally, we study whether SIGN can provide stable long-horizon predictions in challenging settings. In our first study case,
{we simulate coupled FHN dynamics on a partitioned real human-brain structural network with more than \(22{,}000\) nodes for 2,000 time steps under three noise levels: 70~dB, 50~dB, and 30~dB, and predictions are generated by numerically integrating the learned dynamics forward from observed initial conditions.}

{We split the trajectory into a training set (first \(1{,}600\) steps) and a test set (last \(400\) steps). Due to GPU memory constraints, we perform learning and prediction with a fixed prediction horizon of 100 steps. During training, we randomly sample a starting time index \(t\) from the training portion, provide SIGN with the state at that single time point \(x_t\), and roll out the inferred equations for the next 99 steps to obtain \(\{\hat{x}_{t+k}\}_{k=1}^{99}\). The equation parameters are optimized by minimizing the discrepancy between the prediction and the corresponding ground-truth segment \(\{x_{t+k}\}_{k=1}^{99}\). This one-point-to-100-step training protocol discourages autoregressive memorization and instead stresses whether the recovered equations capture a reusable law of motion.}

{For visualization of long-horizon predictions, we further partition the \(2{,}000\)-step trajectory into 20 non-overlapping segments of length 100 (Fig.~\ref{fig41}\textbf{a}). For each segment, SIGN is given only the initial state and then predicts the remaining 99 steps by integrating the inferred equations, yielding a full \(2{,}000\)-step prediction. For baselines (e.g., neural-network predictors), we follow the standard sliding-window evaluation with window length 100 and report the averaged prediction errors over all windows for a fair comparison.}

Across all noise conditions, SIGN generates 
{trajectory predicts that preserve the characteristic slow--fast phase portrait}
(Fig.~\ref{fig41}\textbf{a}).  
{In the 70~dB and 50~dB cases, predicts closely track the ground truth.}
{At 30~dB, where observations are visibly corrupted, predictions remain stable and retain the qualitative FHN geometry.}
{Fig.~\ref{fig41}\textbf{b} summarizes the node-wise prediction error distribution (MSE) under 50~dB and 30~dB noise.}
{Two-dimensional density plots (Fig.~\ref{fig41}\textbf{c}) show that predicted and true values cluster along the identity line, with coefficients of determination \(R^2 = 0.9999\) (50~dB) and \(R^2 = 0.9990\) (30~dB).}
Fig.~\ref{fig41}\textbf{d} reports prediction error across the 20 segments, distinguishing training and prediction intervals. The results at 70 dB can be found in \textcolor{blue}{the SI, Section VI.}

{To connect prediction performance to interpretability, we examine the inferred equations under different noise levels.}
Under noise-free and 50~dB conditions, SIGN identifies the canonical FHN equations, while
under stronger noise (30~dB), SIGN preserves all core nonlinear and coupling terms and introduces 
{only two small sinusoidal corrections,  
\(-0.001\sin(x_1j)\) and \(-0.0071\sin(x_1j - x_1i)\),  
which compensate for noise-induced nonsmooth perturbations.}
{Across noise levels, the recovered equation form remains largely consistent, supporting stable integration-based prediction from noisy observations.}

We benchmark SIGN against five representative graph-based neural predictors, including MTGNN~\cite{wu2020connecting}, ASTGCN~\cite{guo2019attention}, MSTGCN~\cite{guo2019attention}, STSGCN~\cite{song2020spatial}, and STGCN~\cite{yu2017spatio}.  
All baselines were trained on the first \(1{,}600\) steps and asked to predict the next 100-step, using a 100-step input horizon for each sliding-window (Fig.~\ref{fig41}\textbf{e}).  
Despite using drastically less temporal information (one point per segment), SIGN achieves the best prediction accuracy, reaching RMSE values of \(0.0175\) (50~dB) and \(0.0476\) (30~dB), and corresponding MAPE values of \(7.60\%\) and \(30.47\%\).
Black-box models initially fit short-term correlations but exhibit substantial drift during extended predictions, especially under noise.
In contrast, SIGN’s predictions remain coherent because they arise from integrating explicit, noise-tolerant dynamical equations rather than extrapolating from past observations.
These results shows that once stable governing equations are recovered at scale, prediction follows directly via numerical integration, providing long-horizon prediction that remain interpretable even in large, noisy neural systems.

\begin{figure}
    \centering
    \includegraphics[width=1\linewidth]{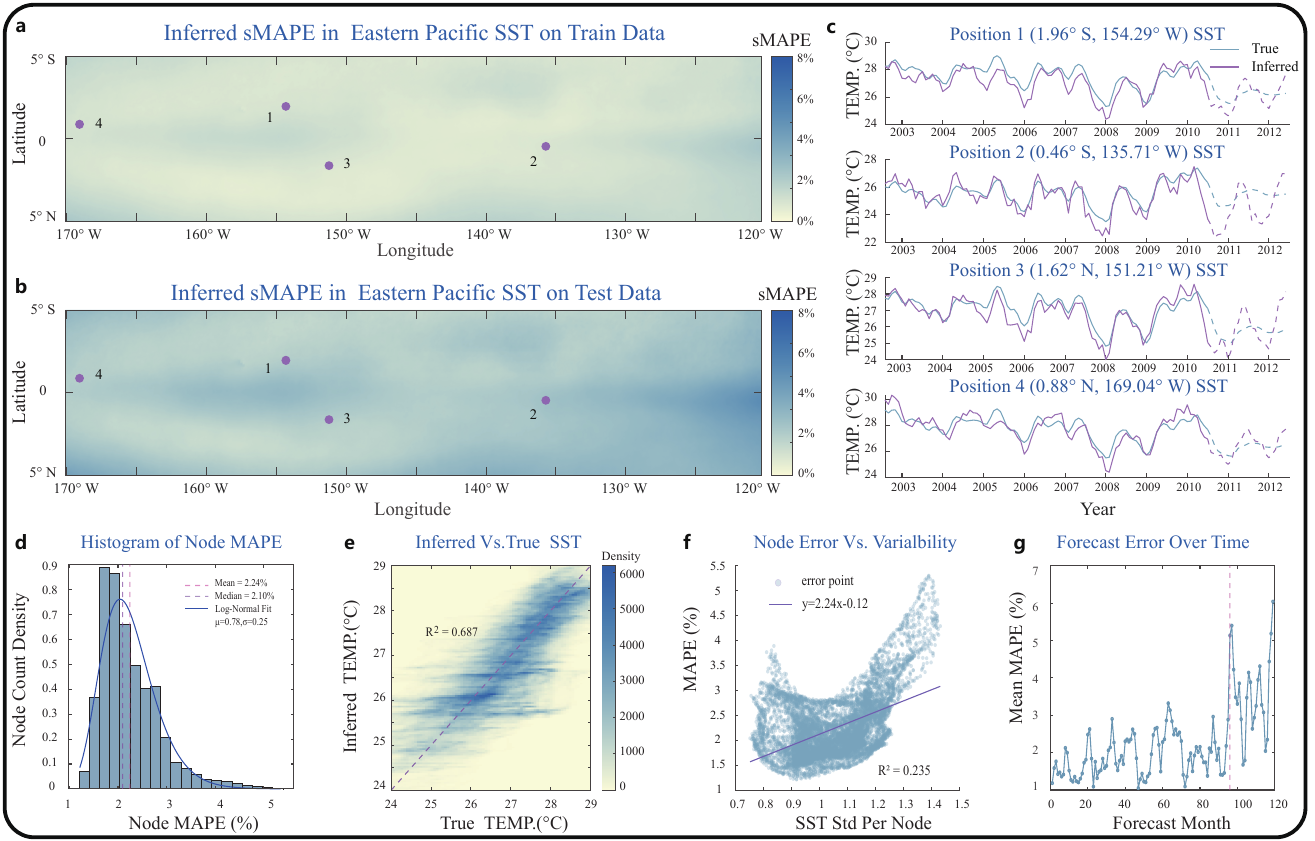}
    \caption{
SIGN equation inference and integration-based prediction of large-scale sea surface temperature (SST) dynamics.
\textbf{a}, Spatial map of node-wise MAPE over the 8-year training period (June 2002--June 2010), summarizing regional prediction error (mean MAPE \(=1.93\%\)).
\textbf{b}, Spatial map of node-wise MAPE over the 2-year testing period (July 2010--June 2012) (mean MAPE \(=3.55\%\)).
\textbf{c}, Representative SST trajectories at four randomly selected grid points: observed SST (blue) versus trajectories obtained by integrating the SIGN-inferred equations (purple). Solid and dashed segments indicate training and testing periods, respectively.
\textbf{d}, Distribution of MAPE aggregated over all nodes and time steps.
\textbf{e}, Two-dimensional density plot comparing predicted versus observed SST values (shown over \(24\)--\(29^{\circ}\mathrm{C}\)); density indicates the frequency of value pairs.
\textbf{f}, Relationship between local SST variability (standard deviation) and node-wise prediction error (MAPE), evaluated on 10{,}000 randomly sampled nodes.
\textbf{g}, MAPE as a function of prediction horizon over the full 10-year window, with the training/testing boundary indicated.
}
    \label{fig4} 
\end{figure}
\subsection{Prediction of climate dynamics}
\label{sec:sstdyn}

In our second study case, we focus on sea surface temperature (SST) dynamics, which provide a stringent real-world test for data-driven discovery.  
The system is high-dimensional, externally forced, noisy, and observed only at coarse monthly resolution.  
Recovering explicit governing equations directly from time series of temperature measurements at different points in space and subsequently using these equations for long-horizon prediction represents a substantial challenge beyond synthetic benchmarks.  
This real-world application directly demonstrates that scalable prediction can be achieved by first inferring a node-invariant governing equation from data.
We focus on a region near the equator in the eastern Pacific Ocean ($5^\circ\mathrm{N}$–$5^\circ\mathrm{S}$, $120^\circ\mathrm{W}$–$170^\circ\mathrm{W}$) using a decade-long SST record (June 2002–June 2012) with spatial resolution $0.083^\circ$ and monthly sampling.  
{We observe sea surface temperature time series at each spatial grid point and treat each grid point as a node. Since we have no direct information about the underlying connectivity, we add undirected edges between nodes within $0.3^\circ$ distance. After removing island points, this yields a graph with 71,987 nodes and 2,281,812 edges.}

As expected, time series of SST exhibit pronounced periodicity due to seasonal and sub-seasonal climate forcing.  
To allow SIGN to capture such temporal structure from data alone, we consider the library of functions that we use for the self-dynamics term $F$ in Eq.~\ref{eq:base} with time-dependent basis functions.  
We compute the Fourier transform of the SST signal averaged across all nodes and extract several dominant frequencies corresponding to annual and intra-seasonal cycles (e.g., periods 1.29, 1.84, 4.3).
For each identified period $t_s$, we include the pair $\sin\left(\frac{2\pi}{t_s}t\right)$ and $\cos\left(\frac{2\pi}{t_s}t\right)$ in the candidate library of $F$.  
This transforms $F(x_i)$ into a time-aware functional form $F(x_i, t)$ which is same for all nodes.  
{This extension introduces no node-specific parameters and preserves the node-invariant assumption while capturing externally driven periodic forcing.}

We split the dataset into an 8-year training period and a 2-year testing period.  
SIGN learns Eq.~\eqref{eq:sst_dyn} solely from the training data, then integrates the inferred dynamics forward to generate the full 10-year trajectory from the initial state.
The spatial distribution of training error (MAPE) is 
{summarized in Fig.~\ref{fig4}\textbf{a}, with mean MAPE \(=1.93\%\) across the region}
(Fig.~\ref{fig4}\textbf{a}); the testing period yields 
{mean MAPE \(=3.55\%\)}
(Fig.~\ref{fig4}\textbf{b}).  
Sample trajectories from four representative sites (Fig.~\ref{fig4}\textbf{c}) show that the inferred dynamics 
{track observations}
, capturing both the fitted period (solid lines) and the 2-year extrapolation (dashed lines), including multi-cycle fluctuations.

The inferred SST dynamics take the form
\begin{equation}
\frac{{\rm d}x_i}{{\rm d}t} = \underbrace{\sum_{s=1}^{S}\!\left(k_s\sin\!\frac{2\pi t}{t_s} + k_{S+s}\cos\!\frac{2\pi t}{t_s}\right) + c\tan(x_i) + C_0}_{\text{Self-dynamics } F(x_i,t)}
\;+\;
\underbrace{\sum_{j\in N_i} \left(a(x_j - x_i) + b\sin(x_j)\right)}_{\text{Coupling dynamics } C}.
\label{eq:sst_dyn}
\end{equation}
{The learned coefficients ($a = 0.4847$, $b = 0.0096$, $c=-0.0798$, $C_0=-0.1334$) indicate a dominant diffusive spatial coupling term, a weak nonlinear correction, and globally shared periodic forcing captured by the Fourier bases. 
All terms are selected from the candidate library directly from observational data.}
Full coefficient details and spectral selection are provided in \textcolor{blue}{the SI, Section~VI}.

{To characterize where prediction is more difficult, Fig.~\ref{fig4}\textbf{d} reports the distribution of MAPE aggregated over nodes and time steps, and Fig.~\ref{fig4}\textbf{e} compares predicted versus observed SST values via a two-dimensional density plot. 
Fig.~\ref{fig4}\textbf{f} relates node-wise SST variability (standard deviation) to node-wise error, and Fig.~\ref{fig4}\textbf{g} reports MAPE as a function of prediction horizon, with the training/testing boundary indicated. Although the test-set MAPE remains low, the corresponding $R^2$ is more modest. This reflects that the model captures the dominant large-scale SST evolution and mean amplitude with relatively small absolute error, while finer local fluctuations and variance on the test period are less fully explained.}

{Together, these results demonstrate that SIGN successfully identifies and predicts large-scale geophysical dynamics, going beyond synthetic benchmarks. It recovers an explicit SST dynamical equation, offering a compact and interpretable foundation for long-term prediction. The high prediction accuracy enables the use of inference-based methods to predict large-scale network dynamics.}
\section{Discussion}

This work introduces SIGN, a scalable symbolic identification framework that reframes ultra-large-scale prediction as a problem of inferring governing equations. 
{Unlike conventional pipelines that separate equation discovery from prediction, SIGN ties prediction directly to recovered dynamics: once an explicit governing equation is inferred, long-horizon prediction follow by numerical integration rather than an auxiliary black-box module.}

Sparse regression methods such as SINDy~\cite{brunton2016discovering} and its network extensions~\cite{casadiego2017model} have established the foundational paradigm of discovering interpretable dynamics from data. 
{However, per-node regression and rapidly growing interaction libraries limit their scalability.}
Recent two-stage and variational extensions~\cite{gao2022autonomous,gao2024learning} alleviate some constraints but still incur computational costs scaling with the number of nodes, 
{which can degrade performance at large scales.}
Neural and graph-based predictors~\cite{wu2020connecting, guo2019attention, yu2017spatio} offer scalability but lack explicit symbolic structure, often requiring large training datasets and exhibiting limited extrapolative power. 
{SIGN occupies a middle ground: a symbolic sparsity prior yields compact, interpretable equations, while GNN-based aggregation decouples inference complexity from network size, bridging interpretability and scalability.}

{Prediction in SIGN follows from learning a globally shared symbolic structure and a fixed-size coefficient set: high-dimensional observations are reduced to a compact mechanistic model, and prediction proceeds by integrating the inferred dynamics rather than extrapolating correlations.}
This explains why SIGN achieves stable long-horizon prediction in challenging regimes such as noisy FitzHugh--Nagumo neural activity and multi-year sea-surface temperature evolution, 
{where purely autoregressive predictors often require dense history and can be sensitive to noise or shift.}

Across six canonical benchmark systems, SIGN consistently recovers the correct governing equations 
{with low symbolic error,}
even on networks with up to $10^5$ nodes. 
Robustness studies further show resilience to noise, low temporal resolution, missing nodes and edges, and even mismatches between the true dynamics and the candidate basis. 
{Recovery on non-canonical systems (e.g., the Chua circuit and fractional mutualistic dynamics) suggests an inductive bias toward parsimonious functional representations.}
In real-world applications, SIGN successfully reconstructed Pacific sea-surface temperature dynamics using a time-augmented basis derived from Fourier modes, producing interpretable equations and accurate two-year prediction from purely observational data.

Several avenues merit further investigation. 
{First, systems with coupled internal states may benefit from multi-channel message passing and cross-variable bases. 
Second, higher-order, delayed, or nonlocal interactions common in biology and climate may require hypergraph or delay-embedded extensions. 
Third, incorporating exogenous drivers and irregular sampling (e.g., via neural ODE tooling~\cite{chen2018neural}) could broaden real-data applicability.}
{By enabling explicit, interpretable, and scalable identification of governing laws in ultra-large networks, SIGN points toward a unified framework in which symbolic discovery directly supports mechanistic prediction, integrating equation-based reasoning with graph neural computation at scale.}

\section{Methods}\label{sce:method}
\noindent
We propose a two-stage framework that enables scalable prediction in large-scale networks by explicitly inferring shared governing equations. 
The framework is built on the assumption that all nodes follow a common intrinsic dynamics and identical coupling mechanisms, allowing the global system to be described by a compact, network-size-invariant set of equations. 
Rather than treating equation discovery and prediction as separate objectives, the proposed formulation decouples the identification of active dynamical terms from parameter estimation, making equation inference tractable at scale while preserving interpretability.  In the first stage, a stable support set is identified via sparse regression combined with clustering. In the second stage, a graph neural network is trained to approximate the global dynamics, constrained to the previously identified support.

\subsection{Stable support mask via DBSCAN consensus}

The first stage addresses the challenge of identifying which basis functions are consistently active across the network. We begin by randomly selecting a small subset of nodes $\mathcal{S} \subset V$, where $|\mathcal{S}| \ll N$, as representative samples. {For each node $s \in \mathcal{S}$, we construct a nonlinear feature library $\theta(x_s(t))$ from its state trajectory $x_s(t)$, and solve a Lasso-type sparse regression problem in discrete time:
\begin{equation}
\xi^{(s)} = \arg\min_{\xi_f, \xi_c} \sum_{t=1}^{T-1} \left\| \dot{x}_s(t) - \xi_f^T \theta_{f}(x_s(t))\,  - \sum_{j=1}^N A_{sj} \left(\xi_c^T \theta_{c}(x_s(t),x_j(t)) \right) \right\|^2 + \lambda \left( \|\xi_f\|_1 + \|\xi_c\|_1 \right),
\end{equation}
where $\dot{x}_s(t)$ denotes the numerical derivative (e.g., finite difference) of the state at time $t$, and $\lambda$ controls the sparsity level. This yields a set of sparse coefficient vectors $\{\xi^{(s)}\}$, each encoding the locally inferred active terms for node $s$.}

{To robustly extract a global support pattern from the locally estimated coefficients \(\{\xi^{(s)}\}\), we apply DBSCAN—a density-based clustering algorithm—on the set of sparse vectors. DBSCAN groups vectors based on pairwise distance using a neighborhood radius \(\epsilon\) and a minimum cluster size \(m_{\text{min}}\). Let $\{\mathcal{C}_1, \mathcal{C}_2, \dots, \mathcal{C}_r\}$ denote the set of valid clusters returned by DBSCAN.
{The global support is computed from the union $\mathcal{C}^{\star} = \bigcup_{i=1}^{r} \mathcal{C}_i$,} assuming all these clusters reflect consistent local dynamics.
To construct a global binary support mask \(M\), we compute the average magnitude of each coefficient index \(k\) across all vectors in \(\mathcal{C}\), and threshold it using an indicator function:
\begin{equation}
M_k = \mathbb{I} \left( {\frac{1}{|\mathcal{C}^{\star}|}} \sum_{s \in \mathcal{C}} |\xi_k^{(s)}| > \delta \right), \quad \text{with } M_k \in \{0,1\},
\end{equation}
where \(\delta\) is a small threshold (typically \(\delta = 0\)) used to identify nonzero entries, and \(\mathbb{I}(\cdot)\) denotes the indicator function. The resulting binary mask \(M\) encodes the consensus set of active basis functions, and serves as a hard sparsity prior for global dynamics inference. This clustering-based consensus strategy filters out spurious terms while preserving the dominant support shared across the most reliable local regressions.}

\subsection{Mask-constrained GNN dynamics learning}
With the binary support mask $M$ fixed, we proceed to train a graph neural network (GNN) that models the global system dynamics while adhering to the identified sparse structure. The evolution of each node state $x_i(t)$ is described by a GNN-based update rule:
\begin{equation}
\hat{x}_i(t+\tau) = x_i(t) + \tau \cdot \mathrm{SIGN}(x_i(t); w),
\end{equation}
where $\mathrm{SIGN}(\cdot)$ is a parameterized function approximated by a GNN, $w$ denotes its learnable parameters, and $\tau$ is the discrete time step. The function $\mathrm{SIGN}$ is decomposed into node-wise and edge-wise contributions over a predefined basis:
\begin{equation}
\mathrm{SIGN}(x_i(t); w) = \sum_{k=1}^{K_f} (M_f[k] \cdot w_f[k])\, \theta_{f}(x_i(t))[k] + \sum_{j \in \mathcal{N}(i)} \sum_{l=1}^{K_c} (M_c[l] \cdot w_c[l])\, \theta_{c}(x_i(t), x_j(t))[l],
\end{equation}
where $\theta_{f}(x_i(t))$ and $ \theta_{c}(x_i(t), x_j(t))$ are basis functions, and $M = [M_f; M_c] \in \{0,1\}^{K_f + K_c}$ specifies the active support. By construction, the parameters are masked as $w = M \odot w$, enforcing hard sparsity during training and inference. The network is trained by minimizing the trajectory prediction error over all nodes and time steps:
\begin{equation}
\min_{w} \sum_{t=1}^{T-\tau} \sum_{i=1}^N \left\| x_i(t+\tau) - \hat{x}_i(t+\tau) \right\|_2^2, \quad \text{subject to } w = M \odot w.
\end{equation}

\subsection{Evaluation metrics}
\label{main:metr}
To assess the accuracy of the inferred coefficients, we adopt the symmetric mean absolute percentage error (sMAPE) \cite{flores1986pragmatic}, defined as:
\begin{equation}
\operatorname{sMAPE} = \frac{1}{m} \sum_{i=1}^m \frac{\left| I_i - R_i \right|}{\left| I_i \right| + \left| R_i \right|},
\end{equation}
where \(m\) denotes the number of terms in the union of the inferred and ground-truth basis functions, and \(I_i\) and \(R_i\) represent the inferred and true coefficients, respectively. The value of sMAPE ranges between 0 and 1, with lower values indicating higher accuracy. Importantly, sMAPE increases not only when coefficient estimates deviate but also when the structural form of the inferred equation differs from the ground truth, thereby capturing both parametric and structural inference errors.

To evaluate the predictive performance of the inferred dynamics, we additionally compute the mean absolute percentage error (MAPE) and the mean squared error (MSE):
\begin{equation}
\operatorname{MAPE} = \frac{1}{T \times N} \sum_{i=1}^{N} \sum_{t=t_0}^{T} \left| \frac{x_i(t) - \hat{x}_i(t)}{x_i(t)} \right| \times 100\%, \quad 
\operatorname{MSE}  = \frac{1}{T \times N} \sum_{i=1}^{N} \sum_{t=t_0}^{T} \left[ x_i(t) - \hat{x}_i(t) \right]^2,
\end{equation}
where \(x_i(t)\) and \(\hat{x}_i(t)\) denote the true and inferred node states at time \(t\), respectively. These metrics quantify the fidelity of the learned dynamics relative to ground truth trajectories.

\section{Data availability}
The simulation networks and data can be generated using the dynamics simulation code available on GitHub. The empirical networks comprise the voles social network \cite{nr-voles}, human and mouse gene regulatory networks \cite{bansal2007infer}, human and fly brain networks \cite{bigbrain}, the GitHub collaboration network \cite{rozemberczki2019multiscale}, and the Catster social network \cite{soc-catster}. All network data is accessible on the Stanford Large Network Dataset Collection (https://snap.stanford.edu/data/) and Network Repository (https://networkrepository.com/). The SST data is aggregated from the SSTG dataset \cite{cao2021new}. 

\section{Code availability}
All the source codes are publicly available at https://github.com/SeuQiShao/sign\_all.

\bibliographystyle{unsrt} 
\bibliography{sn-bibliography}

\section{Acknowledgements}
This research was supported by the Youth Scientist Project of the Ministry of Science and Technology of China (Grant No.~2025YFF0524100), the National Natural Science Foundation of China (Grants No.~62233004,~62273090,~62073076, and~T2541017), the Zhishan Youth Scholar Program of the Southeast University, the Jiangsu Provincial Scientific Research Center of Applied Mathematics (Grant No.~BK20233002), the Basic Research Program of Jiangsu (Grants No.~BK20253018,~BK20253020), the Open Research Project of the State Key Laboratory of Industrial Control Technology, China (Grant No.~ICT2025B54).

\section{Author contributions}
Q.S., D.C. and V.L conceived and designed the research. Q.S., D.C., Y.Z., A.M., W.Y., and W.L. developed the framework and formulated the theoretical model. Q.S., D.C., and J.C. carried out the data research. Q.S., D.C., and J.C. carried out the simulations and analyses. Q.S., D.C., W.Y., V.L. and W.L. wrote the manuscript. All authors contributed to the discussions on the method and the revision of this article. 

\section{Competing interests}
The authors declare no competing interests.
\end{document}